\documentclass[journal]{IEEEtran}
\usepackage[utf8]{inputenc}
\usepackage{amsmath}
\usepackage{amsfonts}
\usepackage{amssymb}
\usepackage{color}
\usepackage{ifpdf}
\usepackage{cite}

\usepackage[pagebackref,hidelinks]{hyperref}
\hypersetup{
    colorlinks,
    citecolor=blue,
    linkcolor=red
}
\usepackage{graphicx}    % for \includegraphics
\usepackage{subcaption}  % for \subfigure
\usepackage{footnote}    % for footnote in table and figure
\usepackage{multirow}    % for table with mutirow
\usepackage{tablefootnote}
\DeclareMathAlphabet\mathbfcal{OMS}{cmsy}{b}{n}

\begin{document}
%%%%%%%%%%%%%%%%%%% ACM MM %%%%%%%%
% Copyright
%\setcopyright{acmcopyright}
%\setcopyright{acmlicensed}
%\setcopyright{rightsretained}
%\setcopyright{usgov}
%\setcopyright{usgovmixed}
%\setcopyright{cagov}
%\setcopyright{cagovmixed}

% DOI
%\doi{}

% ISBN
%\isbn{}

%Conference
%\conferenceinfo{}{}
%\acmPrice{\$15.00}
%\conferenceinfo{WOODSTOCK}{'97 El Paso, Texas USA}
%\CopyrightYear{2007} % Allows default copyright year (20XX) to be over-ridden - IF NEED BE.
%\crdata{0-12345-67-8/90/01}  % Allows default copyright data (0-89791-88-6/97/05) to be over-ridden - IF NEED BE.

%\title{Where to Focus: Query Adaptive Matching for Instance Retrieval Using Convolutional Feature Maps}
%\numberofauthors{1}
%\author{}
%%%%%%%%%%%%%%%%%%%%%%%%%%%%% END %%%%%%%%%%%%%%%

%%%%%%%%%%%%%%%%% IEEE tran %%%%%%%%%%%%%%%%%%%
\title{{\fontsize{23}{24}\selectfont Where to Focus: Query Adaptive Matching for Instance Retrieval Using Convolutional Feature Maps}}
%\title{Where to Focus: Query Adaptive Matching for Instance Retrieval Using Convolutional Feature Maps}

\author{
	{\fontfamily{phv}\fontsize{12}{14.4}\selectfont Jiewei Cao$^1$\thanks{This work was done when Jiewei was visiting The University of Adelaide.}, Lingqiao Liu$^2$, Peng Wang$^1$, Zi Huang$^1$, Chunhua Shen$^2$, and Heng Tao Shen$^1$}
	\vspace{1.6mm}\\
	{\fontfamily{phv}\fontsize{10}{12}\selectfont $^1$The University of Queensland, Australia}
	%\vspace{1mm}\\
	{\fontfamily{phv}\fontsize{10}{12}\selectfont $^2$The University of Adelaide, Australia}
	\vspace{1mm}\\
	{\fontfamily{phv}\fontsize{10}{12}\selectfont \{j.cao3, p.wang6\}@uq.edu.au, \{lingqiao.liu, chunhua.shen\}@adelaide.edu.au, \{huang, shenht\}@itee.uq.edu.au}
}
%%%%%%%%%%%%%%%%%%%%%%%%%%%%% END %%%%%%%%%%%%%%%%%%%%

\maketitle

\begin{abstract}
     Instance retrieval requires one to search for images that contain a particular object within a large corpus. Recent studies show that using image features generated by pooling convolutional layer feature maps (CFMs) of a pretrained convolutional neural network (CNN) leads to promising performance for this task. However, due to the global pooling strategy adopted in those works, the generated image feature is less robust to image clutter and tends to be contaminated by the irrelevant image patterns. In this article, we alleviate this drawback by proposing a novel reranking algorithm using CFMs to refine the retrieval result obtained by existing methods. Our key idea, called query adaptive matching (QAM), is to first represent the CFMs of each image by a set of base regions which can be freely combined into larger regions-of-interest. Then the similarity between the query and a candidate image is measured by the best similarity score that can be attained by comparing the query feature and the feature pooled from a combined region. We show that the above procedure can be cast as an optimization problem and it can be solved efficiently with an off-the-shelf solver. Besides this general framework, we also propose two practical ways to create the base regions. One is based on the property of the CFM and the other one is based on a multi-scale spatial pyramid scheme. Through extensive experiments, we show that our reranking approaches bring substantial performance improvement and by applying them we can outperform the state of the art on several instance retrieval benchmarks.
    
\end{abstract}
%%%%%%%%%%%%%%%%%%%%%%%%%%%%%%%%%%%%%%%%%%%%%%%%%%%%%

\section{Introduction}\label{sec:intro}
%% background
    Instance retrieval is the task of finding a particular instance from a large image corpus.
    In practice, instance retrieval has two major challenges: 1) the large visual appearance deformations due to the object's different positions, poses and illumination when being captured; and 2) various distractors from the natural and less contextual backgrounds.
    In the past decade, a lot of work uses local invariant descriptors \cite{Lowe:sift/2004} to handle instance retrieval, including methods that use large visual codebooks \cite{Sivic:bow/2003,Philbin:oxford5k/2007,Jegou:HE/2010,Mikulik:FineQuantization/2013,Tao:locality/2014} and methods that use compact representations \cite{Perronnin:FV/2010,Jegou:VLAD/2012,Arandjelovic:VLAD/2013,Jegou:Tembed/2014}.
    Although promising performance has been reported on some constrained datasets \cite{Philbin:oxford5k/2007,Philbin:paris6k/2008,Jegou:HE/2008,Arandjelovic:sculptures6k/2011} for these methods, in a more realistic and challenging scenario where the target objects are with cluttered backgrounds, instance-level retrieval still remains a challenging problem \cite{Wang:instre/2015}.
    
    Recent studies show the visual descriptors produced by aggregating the convolutional feature maps (CFMs) from convolutional neural networks (CNNs) achieve state-of-the-art performance for image retrieval \cite{Razavian:CnnINSBaseline/2015,Babenko:spoc/2015,Tolias:rmac/2015,Kalantidis:CroW/2015}.
    Comparing to the conventional local descriptors, these deep learned features capture more semantic information since a CNN is usually trained on large labeled image datasets \cite{ILSVRC15}.
    Most existing methods aggregate the CFMs into a compact global representation for an image, where the query-to-image similarity is evaluated on the image level.
    How to use these deep convolutional features to distinguish the target object from distractors in cluttered backgrounds is still challenging.
    
    In this paper, we follow the direction of CNN-based methods and propose a novel reranking algorithm, namely query adaptive matching (QAM), using CFMs to address the image clutter problem during instance retrieval. The key idea of our method is that instead of relying on a global image representation, we represent each dataset image by a set of base regions. In further, we allow these base regions to freely merge into larger regions. Image features are then extracted from the combined region rather than the whole image. Moreover, we parameterize the regions that can be created via combining the base region and represent the combined feature as a function of a set of parameters. Thus, adjusting these parameters will be equivalent to choosing the region-of-interest to focus for a candidate image. Our method will cast the parameter selection issue as an optimization problem under the objective of maximizing the similarity between the query and the feature obtained from the combined region. The optimal parameters are expected to generate a focus region that mostly covers the object-of-interest.
    
    To implement this idea, we also propose two ways of generating base regions. One is based on the property of the CFM and the other one is based on the multi-scale spatial pyramid. By conducting experiments on the various instance retrieval benchmarks, we show that the proposed QAM reranking, together with the two base region generation approaches can achieve promising results which outperform the state of the art. Besides that, we also discuss several practical issues of using CFMs for instance retrieval, including the choice of deep CNN models and the convolutional layers.
    
\section{Related Work}\label{sec:related_work}
\subsection{Non-CNN Based Retrieval}
    In the past decade, instance retrieval is mainly handled by the methods using local invariant descriptors, such as SIFT \cite{Lowe:sift/2004}.
    Previous work can be roughly divided into two categories:
    1) methods that encode local descriptors into large visual codebooks and sparse representations, namely the Bag-of-Words (BoW) \cite{Sivic:bow/2003,Philbin:oxford5k/2007,Jegou:HE/2010,Mikulik:FineQuantization/2013,Tao:locality/2014};
    and 2) methods that aggregate local descriptors into dense and compact features \cite{Perronnin:FV/2010,Jegou:VLAD/2012,Arandjelovic:VLAD/2013,Jegou:Tembed/2014}.
    Due to the loss of spatial information and the degradation of discriminative power of the descriptor after visual word quantization, BoW models are usually followed by some post-processing steps, e.g., spatial verification \cite{Philbin:oxford5k/2007} or query expansion \cite{Chum:qerevisit/2011}, in order to eliminate false positive results.
    In practice, BoW models adopt an inverted index for efficient search, but the number of images that can be indexed is limited by the search time and index size \cite{Jegou:VLAD/2012}.
    A different strategy is to aggregate the local descriptors into compact representations, e.g., compressed Fisher Vector \cite{Perronnin:FV/2010}, VLAD \cite{Jegou:VLAD/2012,Arandjelovic:VLAD/2013} and T-embedding with democratic aggregation \cite{Jegou:Tembed/2014}.
    However, recent studies \cite{Wan:MMcnnCBIR/2014,Razavian:CnnBenchmark/2014,Babenko:NeuralCodes/2014,Razavian:CnnINSBaseline/2015,Babenko:spoc/2015,Tolias:rmac/2015,Kalantidis:CroW/2015} show that the neuron activations extracted from CNN serve as good image representations, which surpass conventional features in low dimensionality settings.

\subsection{CNN Based Retrieval}
    CNNs are widely used in computer vision since the success of ``AlexNet'' \cite{Krizhevsky:alexnet/2012} in large-scale image classification \cite{ILSVRC15}.
    Recent studies \cite{Wan:MMcnnCBIR/2014,Razavian:CnnBenchmark/2014,Babenko:NeuralCodes/2014} show that the neuron activations of CNNs can be used as generic features for image retrieval, where the features are from the \textit{fully-connected layers}.
    However, these layers are trained on labeled objects to facilitate image classification and hence might not generalize to some instance types. 
    These methods usually require fine-tuning the CNN on the target (or visually similar) datasets \cite{Wan:MMcnnCBIR/2014,Babenko:NeuralCodes/2014} to obtain satisfactory retrieval performance.
    
    Besides of the fully-connected layers, there is an emerging trend \cite{Razavian:CnnINSBaseline/2015,Babenko:spoc/2015,Tolias:rmac/2015,Kalantidis:CroW/2015} toward using the activations of the convolutional layers, named as \textit{convolutional feature maps} (CFMs), as image features which shows superior performance.
    Specifically, Razavian et al. \cite{Razavian:CnnINSBaseline/2015} propose to segment the image into multiple square patches and extract patch descriptors using CNN. During searching, they cross-match all the patches to obtain the best match results. Obviously, this method cannot handle large-scale datasets due to the high computational cost. 
    Babenko et al. \cite{Babenko:spoc/2015} propose a simple but effective CFMs aggregation method based on sum-pooling, which generates compact global representations (256 dimensions) for retrieval. But their performance still lags behind the traditional methods.
    Tolias et al. \cite{Tolias:rmac/2015} propose an aggregation method which first decomposes the CFMs into multiple regions at different scales and then aggregates them via sum-pooling. This method also generates compact representations and outperforms \cite{Babenko:spoc/2015} in most cases.
    Besides that, the authors also propose a new reranking method based on integral max-pooling of CFMs which can roughly locate the target objects.  Their retrieval framework achieves comparable or even better performance compared to traditional methods.
    We use a retrieval pipeline similar to that in \cite{Tolias:rmac/2015}, but we propose a different reranking algorithm which can detect the discriminative regions of the target object more flexibly and precisely, and therefore achieves better retrieval performance.

\section{Instance Retrieval using CFMs}
    As mentioned in previous sections, recent studies \cite{Razavian:CnnINSBaseline/2015,Babenko:spoc/2015,Tolias:rmac/2015,Kalantidis:CroW/2015} have shown that using CFMs as image features achieves promising performance in instance retrieval and our work will follow this direction. Before elaborating our methods, this section will give a brief discussion on CFMs, their properties and their use for instance retrieval. These discussions provide the motivation and inspiration of our methods.
   % In this section, we introduce the structure of CFMs (Sec. \ref{sec:cfm_preliminary}) and reveal the feature patterns inside CFMs (Sec. \ref{sec:cfm_patterns}).
    %Finally, we compare the global and local CFMs aggregation methods for instance retrieval (Sec. \ref{sec:cfm_global_vs_local}).
    
\begin{figure}[!t]
    \centering
    \begin{subfigure}[b]{\linewidth}
        \centering
        \includegraphics[width=\linewidth]{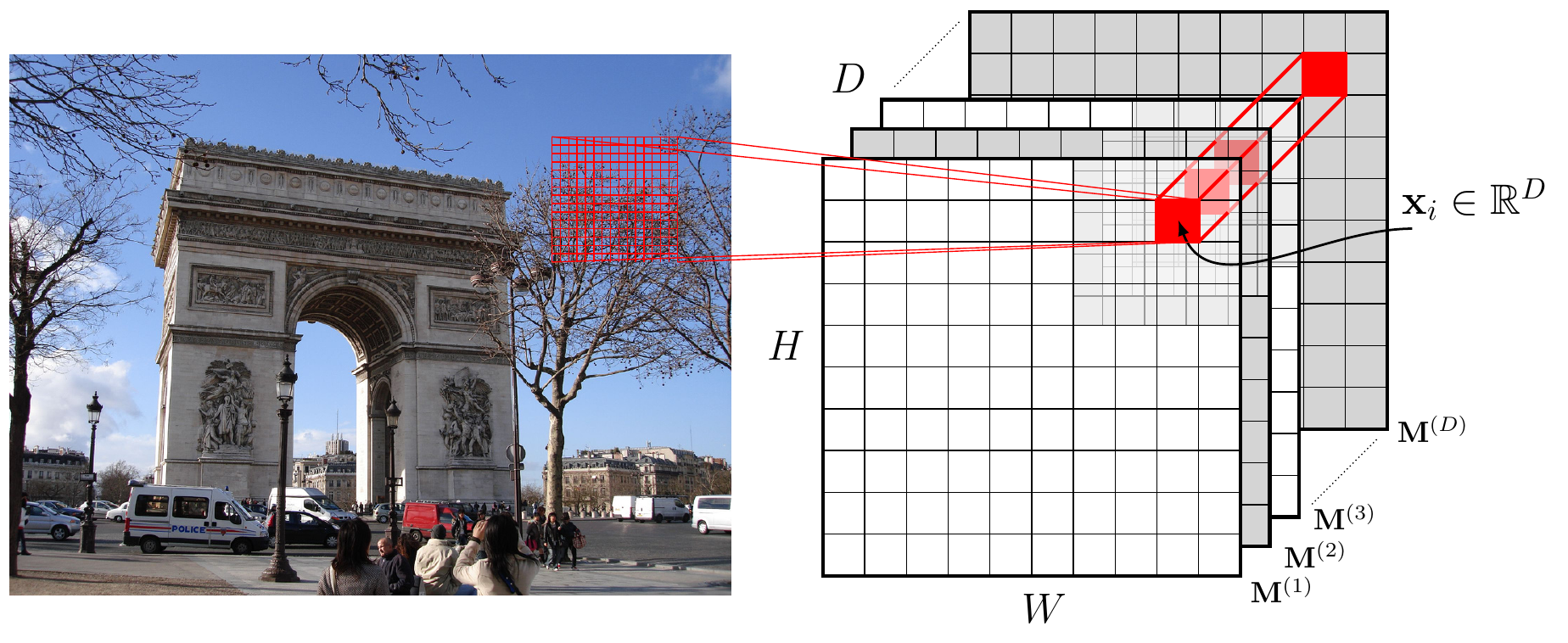}
    \end{subfigure}
    \caption{
    The structure of convolutional feature maps. 
    We feed an image to a pretrained CNN and extract the CFMs from a certain layer, which are consisted of $D$ feature maps with size $H \times W$.
    The neuron activations at each location across all feature maps are composed into a local descriptor (e.g., $\mathbf{x}_i \in \mathbb{R}^D$ at location $i$).
    }
    \label{fig:cfm_structure}
\end{figure}
\subsection{Preliminary}\label{sec:cfm_preliminary}
%% what is CFMs
    The CFMs generated by a convolutional layer can be arranged in a tensor of the size $H \times W \times D$ (see Fig. \ref{fig:cfm_structure}), where $H$ and $W$ denote the height and width of each feature map, and $D$ denotes the number of feature maps (or channels) in that layer.
    Due to the local connectivity of each filter, the activations at the same spatial location across all feature maps can be composed into a $D$-dimensional local descriptor for a certain image region, where the region size equals to the size of the filter's receptive field. This local feature is formally denoted as $\mathbf{x}_i \in \mathbb{R}^D$  in this paper and the subscript $i$ indicates the $i$-th location in the CFMs. In total, there are $H \times W$ such locations in the CFMs. The above arrangement is demonstrated in Fig. \ref{fig:cfm_structure}.
    The major advantage of CFMs over the fully-connected layers' activations is that it retains the spatial information of local image patterns \cite{Long:convCorr/2014,Liu:CrossPooling/2015} and it is essentially more similar to the traditional hand-engineered local features. In fact, the CFMs from one convolutional layer can be viewed as an array of local features sampled from a dense sampling grid.

 %   the local correspondence is retained \cite{Long:convCorr/2014}.
%    This is similar to the conventional hand-engineered features in which local descriptors are computed on small image patches.     
 %   In addition, Mahendran and Vedaldi \cite{Mahendran:convInvert/2015} show that DSIFT and HOG can actually be implemented as CNNs.

\begin{figure}[!t]
        \centering
        \includegraphics[width=\linewidth]{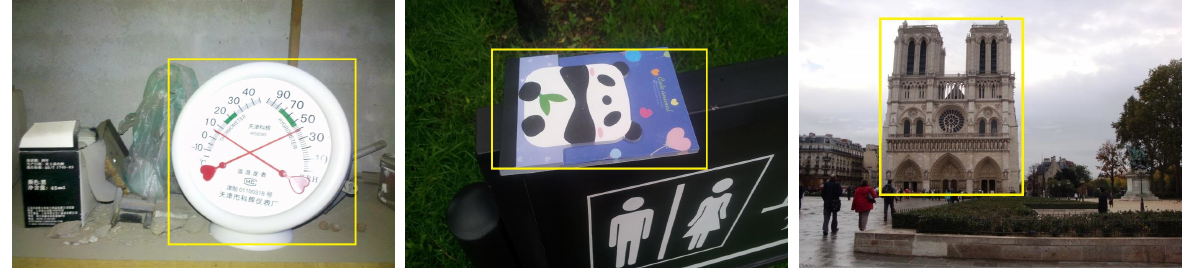}
        \caption{Examples from three different object types in INSTRE  dataset \cite{Wang:instre/2015}: stereoscopic object, planar object and architecture (from left to right). The objects of interest are annotated with yellow bounding boxes.}
        \label{fig:example_imgs}
    \end{figure}

    Throughout this paper, we use the following notations. Let $\mathbf{X} \in \mathbb{R}^{H \times W \times D}$ be the CFMs extracted from the $l$-th layer (after applying the ReLU). It can be equivalently represented as a set of local descriptors $\mathbfcal{X} = \{ \mathbf{x}_i \in \mathbb{R}^D | i \in \{ 1,\dots,H \times W \} \}$. 
    In the following sections, we use three example images selected  from different object types in INSTRE dataset \cite{Wang:instre/2015} for illustration purpose (see Fig. \ref{fig:example_imgs}).
    %where $\mathbf{x}_i \in \mathbb{R}^D$ and $\Omega = \{ 1,\dots,H \times W \}$ is the set of valid 2-D spatial location in $\mathbfcal{X}$.
    %For notation convenience, we also consider the CFMs as a set of feature maps from different channels 
    %$\mathbfcal{X} = \{ \mathbf{M}^{(d)} | d=1,\dots,D \}$, 
    %where $\mathbf{M}^{(d)} \in \mathbb{R}^{H \times W}$.
    %In the following sections, we use three example images selected  from different object types in INSTRE dataset \cite{Wang:instre/2015} for illustration purpose (see Fig. \ref{fig:example_imgs}), and use the pretrained CNN with 19 layers in \cite{Simonyan:vggnet/2014} (named as \textit{VGG19}) to extract CFMs.

\subsection{Understanding CFMs}\label{sec:cfm_patterns}
\begin{figure}[!t]
        \centering
        \includegraphics[width=\linewidth]{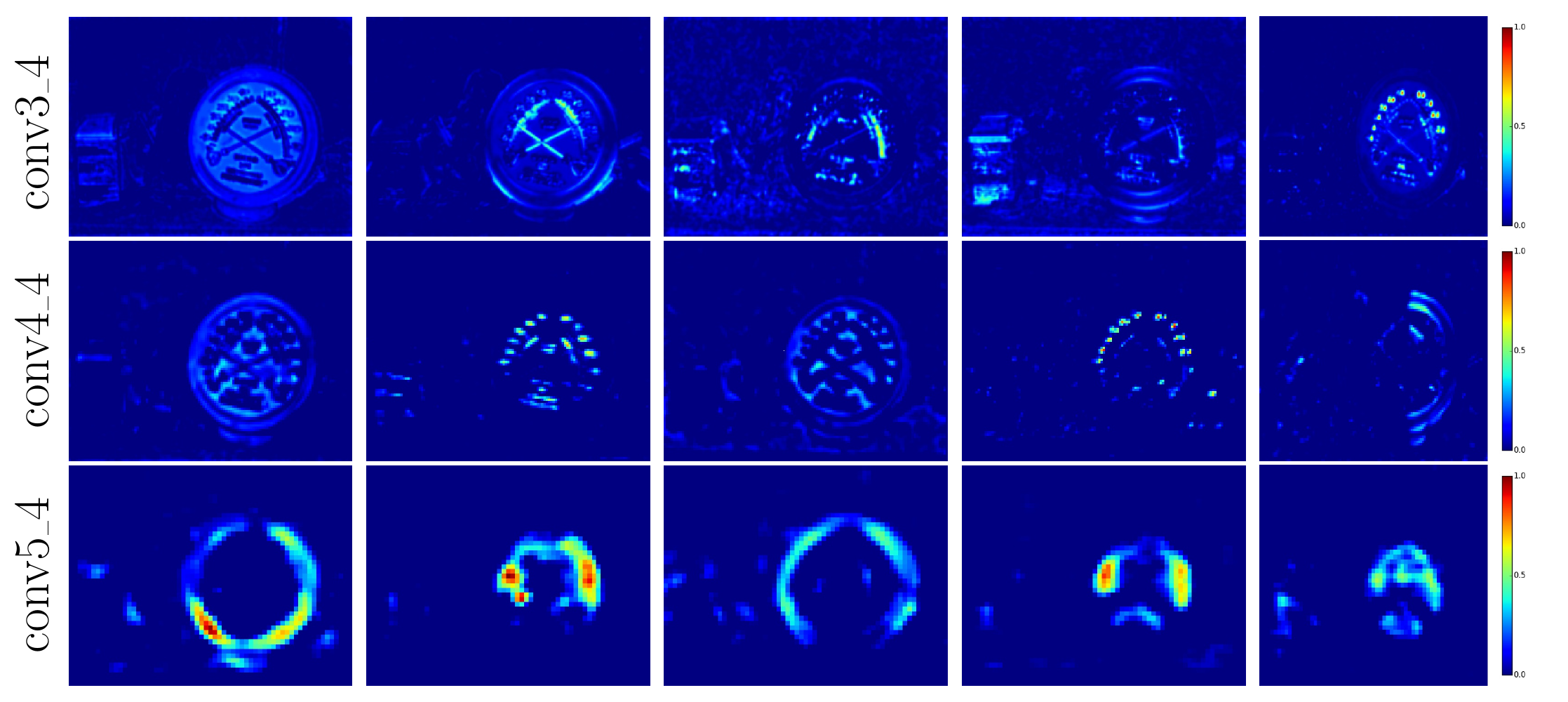}
        \caption{
        The feature maps of the first example image in Fig. \ref{fig:example_imgs} extracted from VGG19 \cite{Simonyan:vggnet/2014}.
        The top five feature maps with the largest average activation score in different layers are selected.
        Each row contains feature maps from a certain layer (\textit{conv3\_4}, \textit{conv4\_4}, and \textit{conv5\_4} respectively).
        %Different columns are the top-5 feature maps.
        }
        \label{fig:cfm_feature_patterns}
    \end{figure}

    Due to the parameter sharing design of the convolutional layer \cite{LeCun:cnn/1989}, a feature map of a convolutional layer can be viewed as the detection scores obtained by applying a detector, the filter of a convolutional layer namely, on $H \times W$ spatial locations and the activation value at the $i$-th location of a CFM characterizes the detector response at the same location. With this analogy, we can obtain an intuitive understanding of a CFM by visualizing it, that is, the highly activated locations suggest that around them it contains the visual patterns depicted in the convolutional layer filter. Fig. \ref{fig:cfm_feature_patterns} gives such an illustration. In Fig. \ref{fig:cfm_feature_patterns}, the feature maps with the top five average activation scores are selected for visualization. Also, comparison of the activation patterns from different convolutional layers is provided in Fig. \ref{fig:cfm_feature_patterns}. From it we can make the following two observations:

%    
%    
%    the hidden neurons in one feature map detect the same feature pattern at different locations.
%    The spatial distributions of different feature patterns provide a discriminative description of the visual content.
%    In order to provide a more intuitive understand, we visualize the feature patterns of CFMs in different layers.
%    Specifically, we calculate the $\ell_1$-norm of each feature map ($||\mathbf{M}^{(d)}||_1$) and select top N feature maps with the largest norms for one layer. These feature maps are then normalized by the maximum value in that layer and visualized.
%    Fig. \ref{fig:cfm_feature_patterns} shows the results of the first example image in Fig. \ref{fig:example_imgs} (Note that similar patterns also exist for other images and hence are not shown here).
%    We observe that:
    \begin{itemize}
    \item A CFM only has high activations on few locations with the presence of certain visual patterns. This suggests that the convolutional filter can be highly selective to certain visual patterns and as a result the CFM can become sparse.
    \item The early convolutional layer of a CNN tends to capture some primary visual patterns, e.g. edges along certain directions or dots while the late convolutional layer is usually selective to visual patterns that corresponding to a shape or an object part, e.g. the circular shape or a wheel region. 
    
    \end{itemize} 
%    For all layers, feature maps in different channels only capture certain feature patterns.
%    The whole image are nicely decomposed into different parts by different feature maps, especially in deeper layers (e.g., \textit{conv5\_4}). 
%    This kind of segmentations is useful for similarity comparison which will be demonstrated in Sec. \ref{sec:cvx:pooling}.
%    \item
%    Most areas in the feature maps are close to zeros, which means the feature maps are vary sparse.
%    Such sparsity also contributes to the discriminative power of CFMs, as will be shown in Sec. \ref{sec:exp_deeper_nets}.
%    \end{itemize}
%    

\subsection{CFMs Aggregation for Instance Retrieval}\label{sec:cfm_global_vs_local}
    To perform instance retrieval, an image needs to be represented by a feature vector or a set of feature vectors. To this end, the CFMs can be aggregated through some pooling strategies which have been explored in several recent works \cite{Razavian:CnnINSBaseline/2015,Babenko:spoc/2015,Tolias:rmac/2015,Kalantidis:CroW/2015}. Most of these works adopt a global pooling strategy, that is, local features from all locations in the CFMs will be aggregated together. In the simplest case, we can adopt the following pooling operations:
    \begin{align}
    	\mathbf{f} = \frac{\sum_i \mathbf{x}_i}{\|\sum_i \mathbf{x}_i\|_2}.
    \end{align}
    In other words, the aggregated image representation $\mathbf{f}$ is obtained by first sum-pooling all the local features and then perform the $\ell_2$-normalization. In practise, the $\ell_2$-normalization has been found to be important and without it the retrieval performance will decrease significantly.
    
    Despite being simple and effective, the global pooling strategy has the drawback of being less robust to the cluttered image background. This is because for instance retrieval the query can be only a cropped object. Thus in a candidate image the object-of-interest may only take up a part of the image whereas global pooling aggregates visual patterns from the whole image area. To alleviate this, one solution is to perform pooling on multiple regions of the image (or equivalently its corresponding CFMs) and create multiple pooled features to represent the image. Then a query will match against all those pooled features and the best matching score will be utilized as the similarity between the query and the image. However, this solution raises the issue of how to choose those regions. On the one hand the number of regions cannot be too large due to the storage bottleneck of a retrieval system. On the other hand the space of all possible sub-regions can be extremely large, especially considering that the object can take any shape and as a result the object shape can be much more complex than a bounding box. Thus, it is very challenging to design a region generation strategy for the multi-region pooling idea.

\section{Query Adaptive CFMs Matching Reranking}\label{sec:cvx}
	 This paper adopts the multi-region pooling strategy discussed in Sec. \ref{sec:cfm_global_vs_local} and proposes an elegant solution to explore a large number of possible regions while keeping the storage for the pooled regions low. Our method first generates a small number of base regions which can be (softly) merged into larger regions. Thus the total number of regions that can be represented by the combination of base regions is very large. For a query image, we compare it against all possible region combinations and pick the best matching score as the query-to-image similarity. At the first glance, the above procedure can be prohibitively expensive. However, if we use a set of parameters to control the merging operation, the pooled feature of the combining region can be formulated as a function of those parameters. The above matching process can be cast as an optimization problem, that is, optimal parameters are sought for maximizing the similarity measure between the pooled feature and the query image feature. The optimal objective value represents the best possible matching score between a combined region and the query image. Certainly, this idea can be slower than direct image feature comparison and thus in this paper we use it as a reranking method and apply it only to a shortlist of images retrieved by an existing approach. In the following subsections, we will elaborate the formulation and implementation details of this idea.

\subsection{Query Adaptive Matching}\label{sec:cvx:opt_sim}
	Formally, our method assumes a set of base image regions have been extracted for each image by a base region generation method, e.g. the ones will be introduced in Sec. \ref{sec:cvx:pooling_fmp} and Sec. \ref{sec:cvx:pooling_ospp}. Let's denote the sum-pooling feature of the CFMs in each of the base region as $\mathbf{f}_k$ and the normalized sum-pooling feature obtained by merging a set of base regions can be represented as\footnote{In practice, $\mathbf{f}_k$ can be generated via other pooling methods, e.g. max-pooling, and we still use Eq. \ref{eq:feature_combine} to combine features from multiple to-be-merged regions. This treatment makes our method more general applicable.}:
	
\begin{align}\label{eq:feature_combine}
	\bar{\mathbf{f}} = \frac{\sum_{k} z_k \mathbf{f}_k}{\|\sum_{k} z_k \mathbf{f}_k\|_2},
\end{align}
where $z_k$ is a binary indicator and if its value equals to 1 its corresponding region will be added into the merged region. In practice, we can also relax its value to a positive real value which implies a soft merge operation. The similarity between the query image $\mathbf{q}$ and merged feature can be then represented by their inner product $\langle \mathbf{q}, \bar{\mathbf{f}} \rangle$. As seen, this similarity is a function of $\{z_k\}$ and our aim is to find the best similarity score that can be achieved by softly merging the base regions. Then this best match score will be used to rerank the shortlist. The search for the optimal $\{z_k\}$ can be written as the following optimization problem:
%    We formalize the similarity calculation between the query and the candidate image as an optimization problem whose goal is to maximize the similarity between the query feature $\mathbf{q}$ and the aggregated region descriptors $\sum^K_{i=1} z_i \mathbf{f}_i$ as shown in Eq. \ref{eq:cvx:max_simple}:
 \begin{equation}
        \begin{aligned}
        & \max  
        && \frac{  \langle \mathbf{q}, \sum^K\limits_{k=1} z_k \mathbf{f}_k\rangle }{ || \sum^K\limits_{i=1} z_k \mathbf{f}_k ||_2 }.\\
        & \text{s.t.} 
        && z_k \geq 0, \forall k=1,...,K.
        \end{aligned}
        \label{eq:cvx:max_simple}
 \end{equation}
  For notation simplification,  Eq. \ref{eq:cvx:max_simple} can be expressed in a more compact way:
    \begin{equation}
        \begin{aligned}
        & \max\limits_{\mathbf{z}}  
        && \frac{ \mathbf{q}^T \mathbf{F} \mathbf{z} }{ \|\mathbf{F} \mathbf{z}\|_2 },\\
        & \text{s.t.} 
        && \mathbf{z} \geq 0.
        \end{aligned}
     \label{eq:cvx:max_matrix}
    \end{equation}
    where $\mathbf{z} = [z_1,  \cdots z_k, \cdots, z_K]^T$ and $\mathbf{F} = [\mathbf{f}_1, \cdots,\mathbf{f}_k, \cdots \mathbf{f}_K]$. 
    %We try to find out the optimal aggregation weighting $\mathbf{z}$ for region descriptors $\mathbfcal{F}$ which can maximize the similarity between the query and the candidate image.
    %For candidates that contain the target object, this method will maximize their similarity scores and hence produce better ranking results.
    For $\mathbf{q}^T \mathbf{F} \mathbf{z} \neq 0$, solving Eq. \ref{eq:cvx:max_matrix} is equivalent to solving:
    \begin{equation}
        \begin{aligned}
        & \min\limits_{\mathbf{z}} 
        && \frac{ \|\mathbf{F} \mathbf{z}\|_2 }{ \mathbf{q}^T \mathbf{F} \mathbf{z} },\\
        & \text{s.t.} 
        && \mathbf{z} \geq 0.
        \end{aligned}
        \label{eq:cvx:min_max_transform}
    \end{equation}
    We further constrain $\mathbf{q}^T \mathbf{F} \mathbf{z} =1$ to avoid the arbitrary scaling of $\mathbf{z}$ and thus ensure that the optimization problem has the unique solution. The final optimization problem is as follows:
    \begin{equation}
        \begin{aligned}
        & \min\limits_{\mathbf{z}} 
        && \|\mathbf{F} \mathbf{z}\|^2_2,\\
        & \text{s.t.} 
        && \mathbf{q}^T \mathbf{F} \mathbf{z} = 1,\\
        &&& \mathbf{z} \geq 0.
        \end{aligned}
        \label{eq:cvx:min_final}
    \end{equation}
    Eq. \ref{eq:cvx:min_final} is a quadratic programming (QP) problem which can be solved efficiently by existing optimization packages. We use the inverse of this optimal value, which equals to the objective value in Eq. \ref{eq:cvx:max_simple}, as the image similarity score.

\subsection{Base Region Generation}\label{sec:cvx:pooling}
The above method applies to base regions generated in any possible ways. In this work, we particularly focus on two approaches to create such base regions.
\subsubsection{Feature Map Pooling (FMP)}\label{sec:cvx:pooling_fmp}
\begin{figure}[!t]
        \centering
        \includegraphics[width=.8\linewidth]{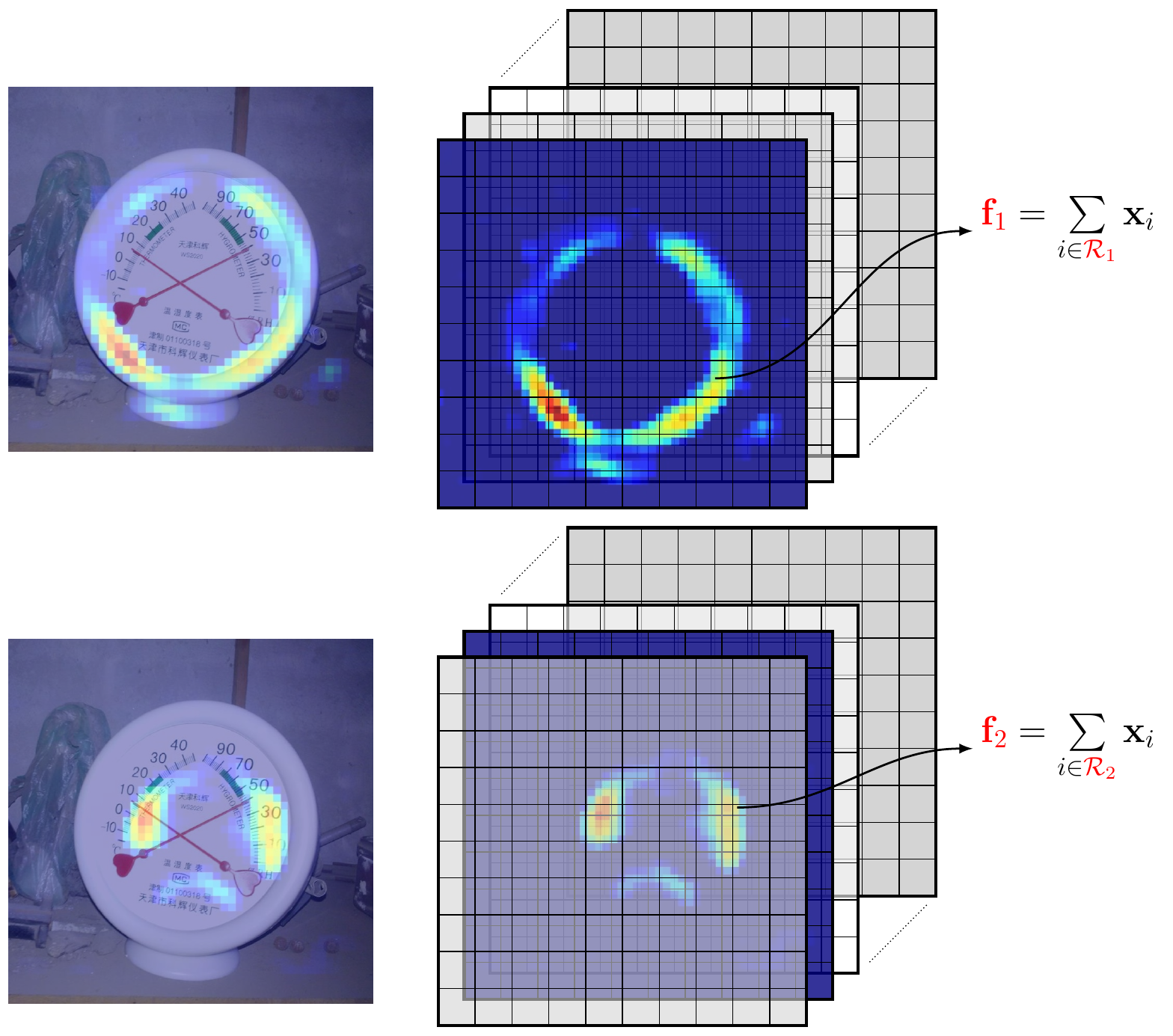}
        \caption{
        Illustration of Feature Map Pooling. 
        For each feature map, the local descriptors within activated regions (warmer parts) are aggregated.
        Essentially, each feature map is treated as a binary mask to select which local descriptors should be aggregated.}
        \label{fig:cvx:fmp}
    \end{figure}

	Our first way of generating base regions is inspired by the observation made in Sec. \ref{sec:cfm_patterns}. As discussed in Sec. \ref{sec:cfm_patterns}, the activated region of a CFM at a late convolutional layer usually represents an object part. The union of them can then cover the whole image or at least those regions with meaningful visual patterns. Thus, the activated region of each CFM in a convolutional layer can be directly utilized as a base region. Specifically, given a set of CFMs $\mathbfcal{X} \in \mathbb{R}^{H \times W \times D}$, we define the set of non-zero elements' locations in the $d$-th feature map as a base region, that is, $\mathcal{R}_{d}=\{ i | x^{d}_i > 0 \}$. The pooled feature for this base region is then calculated as:
    \begin{equation}
    		\mathbf{f}_d = \sum\limits_{ i \in \mathcal{R}_{d} } \mathbf{x}_i.
    \end{equation}
    In our implementation, we also have $\mathbf{f}_d$  $\ell_2$-normalized to compensate the region size discrepancy. 
    We call this base region generation method as Feature Map Pooling, or FMP in short. With FMP, the $\mathbfcal{X}$ will be transformed into $D$ region descriptors which may be corresponding to different object parts, as shown in Fig. \ref{fig:cvx:fmp}.
    
    % \begin{figure}[!t]
    %     \centering
    %     \includegraphics[width=.8\linewidth]{fig/fmp}
    %     \caption{Illustration of Feature Map Pooling. The colored area in each feature map denotes the non-zero elements.}
    %     \label{fig:cvx:fmp}
    % \end{figure} 
    
	Once a convolutional layer of a pretrained CNN is chosen, the number of base regions will equal the number of feature maps or channels in that convolutional layer. However, for a given image, the activated regions for many CFMs can be highly overlapped. Thus it would be a waste of computation if we use all the base regions generated from the CFMs. To reduce the computational cost, in this paper we merge base regions as follows. We first represent each base region by their corresponding pooled feature $\mathbf{f}_d$. If two CFMs have similar activated regions, their corresponding pooled feature will be similar. Thus we run a clustering method, $k$-means clustering or spectral clustering \cite{Luxburg:sc/2007}, to group $D$ pooled features into $K$ clusters. The base regions whose pooled features are grouped together will then be merged into one base region. In our experiment, this merging operation reduces the number of base regions from 512 to around 30 and thus significantly reduces the computational cost.
   
\subsubsection{Overlapped Spatial Pyramid Pooling (OSPP) }\label{sec:cvx:pooling_ospp}
\begin{figure}[!t]
        \centering
        \includegraphics[width=.5\linewidth]{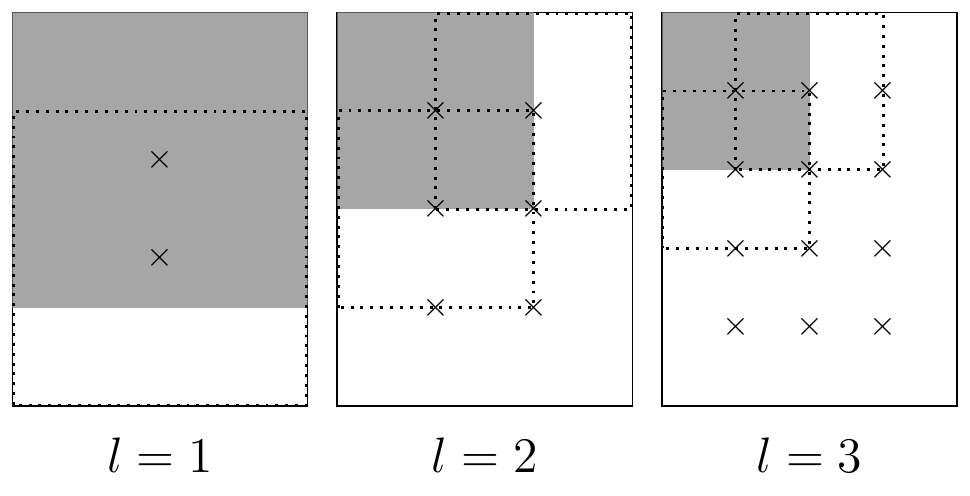}
        \caption{Illustration of region sampling at different scales from \cite{Tolias:rmac/2015}. We show the top-left region of each scale (gray colored region) and its neighboring regions towards each direction (dotted borders). The center of each region is denoted with a cross.}
        \label{fig:cvx:illustration_rmac}
    \end{figure}

	Another way to generate base regions is to directly sample some rectangle regions from the image. Tolias et al. \cite{Tolias:rmac/2015} propose an region sampling method to sample regions at different scales and different locations. In our work, we adopt similar strategy to generate the base regions. More specifically, as shown in Fig. \ref{fig:cvx:illustration_rmac},  we sample square regions at $L$ different scales.  Given an image's CFMs $\mathbf{X} \in \mathbb{R}^{H \times W \times D}$, we uniformly sample regions of width $2 min(W,H)/(l+1)$ at the $l$-th level.  Also, the regions are sampled to allows around 40\% overlap between consecutive regions.
	
	Note that even using a similar sampling strategy, our approach is different from the work in\cite{Tolias:rmac/2015}. The latter one pools the features generated from each sampled region together to obtain a single global-level image representation while our method uses sampled regions as base regions which are fed into the proposed query adaptive matching algorithm.
    %Finally, the local descriptors in each region are pooled and post-processed with $\ell_2$-normalization, PCA-whitening and $\ell_2$-normalization.
    
    \subsection{Retrieval Pipeline}\label{sec:cvx:pipeline}
Finally, to achieve the best performance with our method, we adopt the following pipeline to perform retrieval. 
\begin{itemize}
\item 
    \textbf{Initial Retrieval}. 
    In offline pre-processing, the CFMs are extracted for all database images via pretrained CNNs.
    The global image representations are computed by an existing aggregation method (e.g., SPoC \cite{Babenko:spoc/2015}, R-MAC \cite{Tolias:rmac/2015} or CroW \cite{Kalantidis:CroW/2015}).
    During online searching, we extract the same feature for query image and evaluate the similarities between the query and all the database images by calculating the feature similarity, e.g., inner product, to obtain an initial ranking list.
    
\item 
    \textbf{Reranking}. 
    A short list consisting of the top N candidates from the initial ranking list is selected for reranking.  Base region descriptors are extracted offline and now they are used to represent each candidate images.  We solve Eq. \ref{eq:cvx:min_final} and obtain the similarity score between the query and a candidate image for reranking.

\item 
    \textbf{Query Expansion} (QE).
    Finally, we select the top five images after reranking as in \cite{Tolias:rmac/2015} and aggregate their global representations with the query feature by average pooling. We use the aggregated feature as a new query and perform the retrieval again to obtain the final ranking list.

\end{itemize}

\section{Experiments}\label{sec:exp}
\subsection{Experiment Settings}
\subsubsection{Datasets}
    We evaluate our methods on the following datasets:
    \begin{itemize}
        \item 
        \textbf{Oxford5k} \cite{Philbin:oxford5k/2007} contains 5,062 images corresponding to 11 Oxford landmarks.
        There are five possible queries for each landmark, and 55 queries in total.
        This dataset can be combined with the 100,071 distractors from \textit{Flickr100k} \cite{Philbin:oxford5k/2007} for large-scale retrieval evaluation. We denote the combined dataset as \textit{Oxford105k}.
        
        \item 
        \textbf{Paris6k} \cite{Philbin:paris6k/2008} is similar to Oxford5k but consists of 6,412 images corresponding to 12 Paris landmarks. It can also be mixed with the Flickr100k, \textit{Paris106k} namely, for large-scale retrieval evaluation.
        
        \item
        \textbf{Sculptures6k} \cite{Arandjelovic:sculptures6k/2011} has 6,340 smooth sculpture images, which are different from the previous two landmark datasets.
        The whole corpus is divided equally into training and testing datasets.
        Since our focus is generic instance retrieval, we evaluate our methods without using any training images for CNN learning or fine-tuning.
        There are 70 queries in total.
                
        \item 
        \textbf{INSTRE} \cite{Wang:instre/2015} includes 23,070 images from 200 objects and one million distractor images.
        These objects can be categorized into three classes: architectures (buildings and sculptures), planar objects (designs, paintings and planar surface) and daily stereoscopic objects (toys and irregularly-shaped products).
        This dataset is challenging since there are diverse variations for each object and most of the objects are with cluttered and less contextual backgrounds.
        All the annotated images (23,070 in total) from the 200 objects are used as queries for performance evaluation purpose.
    \end{itemize}
    The query bounding boxes are provided on all the datasets and are used for cropping out the target objects during retrieval.
    The evaluation metric is mean average precision (mAP) for all the datasets.

\subsubsection{CFMs Extraction}
    We use VGG19 \cite{Simonyan:vggnet/2014} provided by Caffe \cite{Jia:caffe/2014} and extract the CFMs from the last convolutional layer (\textit{conv5\_4}) because previous studies \cite{Babenko:spoc/2015,Tolias:rmac/2015,Kalantidis:CroW/2015} show that this layer achieves the best performance.
    Moreover, the original image sizes are kept for both database images and queries during feature extraction.
    However, for the large images (whose areas are greater than $1000 \times 1000$) and the small images (whose shorter sides are smaller than 256), we resize them accordingly but keep their original aspect ratios.
    All the input images are zero-centered by \textit{RGB mean pixel subtraction} \cite{Iandola:DenseNet/2014}. 
    
\subsubsection{Global Feature Generation}\label{sec:exp:settings_global_features}
    We evaluate two state-of-the-art CFMs aggregation methods: the sum-pooled convolutional features (SPoC) without center prior \cite{Babenko:spoc/2015} and the regional maximum activation of convolutions (R-MAC) \cite{Tolias:rmac/2015}.
    %The procedure of SPoC is first to aggregate the local descriptors of CFMs into a $D$-dimensional descriptor by sum-pooling, and then perform $\ell_2$-normalization, PCA-whitening and $\ell_2$-normalization at last.
    %The procedure of R-MAC is first to sample a number of regions at different locations and scales from CFMs (as shown in Fig. \ref{fig:cvx:illustration_rmac}), and the local descriptors in each region are max-pooled in to a $D$-dimensional region descriptor. Similarly, the region descriptors are post-processed with $\ell_2$-normalization, PCA-whitening and $\ell_2$-normalization. Finally, all the region descriptors are combined into a single image representation by sum-pooling and $\ell_2$-normalizing at last.
    We use the optimal settings reported in \cite{Babenko:spoc/2015} and \cite{Tolias:rmac/2015}.
    Both SPoC and R-MAC require PCA-whitening and the parameters of PCA are learned on hold-out images for all datasets \cite{Babenko:spoc/2015}. 
    Specifically, when performing retrieval on Oxford5k we learn the parameters on the images from Paris6, and vice versa. For Sculptures6k and INSTRE, we learn the parameters on Flickr100k.
    %Note that PCA-whitening is applied but we keep the original dimensionality.
    In the following experiments, we use the \textit{inner product} as similarity measurement during retrieval for both SPoC and R-MAC. 
    All the experiments are conducted on a Linux server with 40 3.0GHz CPU cores and 256GB memories.

\subsection{Performance of QAM Reranking}\label{sec:exp:reranking}
    We first report the initial retrieval results of the aforementioned global representations in Tab. \ref{tab:exp:map_vgg19}, which serve as the baselines of the proposed method.
    Note that the mAPs of SPoC reported here are higher than those reported in \cite{Babenko:spoc/2015} (e.g., the mAPs are increased from 0.531 to 0.666 on Oxford5k, and 0.501 to 0.606 on Oxford105k).
    This is due to the implementation differences: 1) in our implementation the image is not resized to square shape as in \cite{Babenko:spoc/2015} when being fed into a CNN for feature extraction; 2) the original dimensionality of the representation is kept during the PCA-whitening step.
    From the table we can see R-MAC outperforms SPoC on all the datasets except on Sculptures6k. Therefore, we use R-MAC for initial retrieval when evaluating the performance of our proposed QAM reranking. 
    %Next, we evaluate the performance of QAM reranking on the top 100 images.
    %Next, we evaluate the performance of QAM reranking on two types of region descriptors: the feature map pooling (FMP) descriptors in Sec. \ref{sec:cvx:pooling_fmp} and the overlapped spatial pyramid pooling (OSPP) descriptors in Sec. \ref{sec:cvx:pooling_ospp}.
    %Unless specific, we re-rank the top 100 images.
\begin{savenotes}
\begin{table}[!t]
\centering
\caption{The mAPs of VGG19 with SPoC and R-MAC. ``DIM.'' denotes the feature dimensionality.}
\label{tab:exp:map_vgg19}
\scalebox{0.9}{
\begin{tabular}{l|c|cccccc}
    \hline
                                             & DIM.  & Ox5k  & Ox105k & Pa6k  & Pa106k & Sc6k  & INSTRE \\ \hline
    SPoC          
                                            & 512  & 0.685  &0.622  &0.798  &0.704  &0.517  &0.306 \\  
    R-MAC
                                            & 512  & 0.695 & 0.644 & 0.838  &0.763  &0.490  &0.415  \\                                              
    \hline
\end{tabular}
}
\end{table}
\end{savenotes}

\subsubsection{QAM with FMP}
    QAM is based on the query feature $\mathbf{q}$ and the region descriptors $\mathbf{F}$ of a candidate image.
    To obtain $\mathbf{q}$, the query's CFMs are sum-pooled and $\ell_2$-normalized.
    For database images, we follow the procedure described in Sec. \ref{sec:cvx:pooling_fmp} to obtain their region descriptors.
    % , where each image is represented by 512 region descriptors $\mathbf{F} \in \mathbb{R}^{512 \times 512}$ when using VGG19.
    The results are shown in Tab.~\ref{tab:fmp_different_k}. Here we choose R-MAC for initial retrieval, the results of which serve as the baseline of our reranking method. From the results we see improvement of the reranking over the initial retrieval on all the datasets. 

\begin{figure*}
    \centering
    \includegraphics[width=\linewidth]{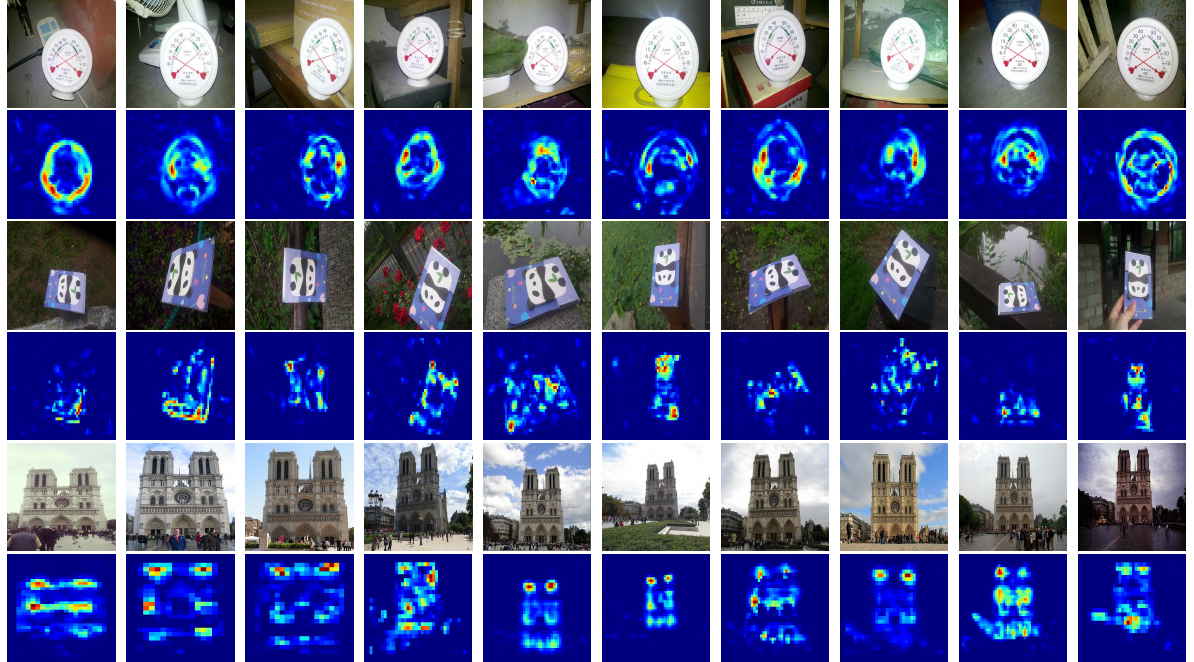}
    \caption{
    Visualization of merged feature maps after QAM.
    The three example images in Fig. \ref{fig:example_imgs} are used as queries.
    The first row in each group displays the top 10 ranked images after QAM reranking with FMP, and the second row shows their merged feature maps.
    }
    \label{fig:exp:rerank_vis}
\end{figure*}

    To demonstrate the effectiveness of QAM with FMP more intuitively, we visualize some merged feature maps after QAM in Fig.~\ref{fig:exp:rerank_vis}. 
    We use the three example instances in Fig.~\ref{fig:example_imgs} as queries and visualize the top 10 ranked candidates after reranking. 
    Recall that the objective of QAM is to find the optimal merging of base regions of candidate images which maximizes the similarity to the query feature.
    After solving the matching problem in Eq.~\ref{eq:cvx:min_final}, we obtain the optimal aggregation weighting $\mathbf{z}$.
    We multiply the $i$-th feature map by its corresponding weighting score $z_i$, and then sum up all the feature maps to form a merged map.
    The locations with large values (the warm areas) are selected by QAM as the discriminative parts w.r.t the query, and the locations with zero values (the blue areas) tend to correspond to the irrelevant background which are ignored by the QAM.
    As seen, QAM detects the discriminative components of the target object and simultaneously suppresses the distractors.
    Another advantage of QAM with FMP is that it can select object parts in irregular shapes which can largely benefit feature aggregation.  
    
% \begin{figure}[!t]
%     \centering
%     \begin{subfigure}[b]{\linewidth}
%         \includegraphics[width=\linewidth]{fig/vis_rerank_img1}
%     \end{subfigure}
%     \begin{subfigure}[b]{\linewidth}
%         \includegraphics[width=\linewidth]{fig/vis_rerank_img2}
%     \end{subfigure}
%     \begin{subfigure}[b]{\linewidth}
%         \includegraphics[width=\linewidth]{fig/vis_rerank_img3}
%     \end{subfigure}
%     \caption{
%     Visualization of merged feature maps after QAM.
%     The three example images in Fig. \ref{fig:example_imgs} are used as queries.
%     The first row in each group displays the top five ranked images after QAM reranking with FMP, and the second row shows their merged feature maps.
%     }
%     \label{fig:exp:rerank_vis}
% \end{figure}

     As mentioned in Sec.~\ref{sec:cvx:pooling_fmp}, FMP generates 512 base regions for each database image among which some are overlapped. To reduce computational cost, our treatment is to perform spectral clustering to pre-merge the regions offline and the number of clusters correspond to the number of regions after this merging. To obtain a trade-off between reranking performance and retrieval efficiency we test the performance of FMP under different number of clusters $K$ in Tab \ref{tab:fmp_different_k}.
    As can be seen, initially increasing the number of clusters leads to better performance. One interesting discovery is that when $K=25$, the performance is comparable to or even better than that obtained without clustering. And after $K=25$, the results remain stable when continuously increasing $K$.
    Based on this observation, we fix $K=25$ in the following experiments.
    With this pre-merging step, we downsize the FMP features from $\mathbb{R}^{512 \times 512}$ to $\mathbb{R}^{25 \times 512}$.
    
    % However, in practice, 512 region descriptors for each image costs too much memory footprints. 
    % As discussed in Sec. \ref{sec:cvx:pooling_fmp}, we apply spectral clustering to merge them into $K$ descriptors.
    % Specifically, to obtain an optimal trade-off between reranking performance and retrieval efficiency we test the performance of FMP under different number of clusters $K$ in Tab \ref{tab:fmp_different_k}.
    % As can be seen, initially increasing the number of clusters can lead to better performance. But when $K$ reaches 25, the results remain stable.
    % The merged region descriptors achieve competitive or even better results.
    % Based on this observation, we set $K=25$ in the following experiments because it achieves comparable performance compared to $K=30$ but requires $16\%$ less memory footprints.
    % With this setting, we downsize the FMP features from $\mathbb{R}^{512 \times 512}$ to $\mathbb{R}^{25 \times 512}$.

\begin{table}[!t]
\centering
\caption{
    The mAPs before and after reranking with FMP.
    ``BEF.'' means the R-MAC results (in Tab. \ref{tab:exp:map_vgg19}) before QAM reranking.
    ``AFT.'' reports the reranking performance of original FMP descriptors before merging, and the following columns report the performance of compressed FMP descriptors with different cluster numbers.
    Note that we stop testing $K$ on large datasets (bottom row).
}
\label{tab:fmp_different_k}
\scalebox{0.9}{
\begin{tabular}{l|l|l|llll}
\hline
       & BEF.  & AFT.  & $K=10$ & $K=20$ & $K=25$ & $K=30$                    \\ 
\hline
Ox5k   & 0.695 & 0.709 & 0.689 & 0.706 & \textbf{0.710} & 0.709               \\
Pa6k   & 0.838 & 0.844 & 0.844 & 0.844 & \textbf{0.845} & \textbf{0.845}      \\
Sc6k   & 0.490 & \textbf{0.515} & 0.512 & 0.513 & 0.514 & \textbf{0.515}      \\ 

\hline
Ox105k & 0.644 & 0.658 & - & - & \textbf{0.662} & - \\ 
Pa106k & 0.763 & \textbf{0.774} & - & - & \textbf{0.774} & - \\
INSTRE & 0.415 & 0.436 & - & - & \textbf{0.446} & - \\

\hline    
\end{tabular}
}
\end{table}

\subsubsection{QAM with OSPP}
    In this subsection, we evaluate the QAM reranking with the other base region generation strategy overlapped spatial pyramid pooling (OSPP).
    When reranking with OSPP, the query's CFMs are aggregated via R-MAC and for candidate database images, we follow the procedure described in Sec. \ref{sec:cvx:pooling_ospp} to generate their region descriptors.
    The number of scales is set to 3 ($L=3$) for both query and candidates which is the same as in \cite{Tolias:rmac/2015}.
    The reranking results are show in Tab. \ref{tab:ospp_map}. As can be seen, the reranking with OSPP consistently outperforms the initial retrieval on all the datasets, which demonstrates the effectiveness of the proposed reranking method.

    Moreover, we draw a comparison between the two proposed reranking methods in Tab.~\ref{tab:ospp_map}. Based on different region generation mechanisms, these two reranking methods have different advantages. For FMP, it can generate base regions that correspond to object parts of irregular shapes which are more flexible than regular shapes such as rectangles. For OSPP, it considers the multi-scale information which makes the matching more robust. Thus, we see these two reranking strategies have different performances on the datasets. But generally, their performance are comparable and both outperform the R-MAC based initial retrieval. 
    %The QAM with OSPP achieves competitive results compared to FMP: they outperform each other on Oxford5k and Sculptures6k.
    % Both reranking methods improve the retrieval performance and the results of FMP are better except for Oxford5k.
    % The reason might be the base regions generated by FMP have irregular shapes and are therefore better at separating the objects from cluttered backgrounds, while the base regions of OSPP are just some square segments at different scales and locations. 
    % Nevertheless, OSPP achieves competitive results compared to FMP.

\subsubsection{Computational Cost of QAM Reranking}
    Since efficiency is an important aspect of instance retrieval, in this subsection, we evaluate both the storage cost and computational cost of the proposed reranking strategies. Specifically, we report the average cost of performing one similarity matching. The results are shown in Tab.~\ref{tab:exp:rerank_costs_vgg19}. For FMP based reranking, we use clustering to merge the region number of a database image from 512 to 25 offline which significantly reduces the memory footprints and the computational cost of QAM. For OSPP, the number of regions of a database image is 23 on average. The results demonstrate that both QAM strategies can be performed efficiently, about 15 ms, under tractable memory footprints. Since the reranking is conducted on the top 100 shortlist of initial retrieval, it costs about 1.6s for each query.
    
%     We further compare the computational costs of FMP and OSPP as shown in Tab. \ref{tab:exp:rerank_costs_vgg19}.
%     Since the last convolutional layers of VGG19 has 512 channels, the dimensionality of each region descriptor is 512. 
%     We use single-precision floating point numbers to store the features.
% %    Due to the larger memory footprints and longer computation time required by FMP, we choose the QAM reranking with OSPP descriptors as our optimal reranking settings. However, for applications where precision is crucial, FMP should be preferred.
%     Overall, the memory and time consumption is similar for both types of multi-region features.
%     One drawback of FMP is that it requires spectral clustering to reduce the feature dimensionality for each image and hence takes longer time to generate these features. 
%     However, this does not affects the online search time as the multi-region features are generated offline.
%     Therefore, we choose the QAM reranking with FMP descriptors as our optimal reranking settings.

\begin{table}[!t]
\centering
\caption{
    The mAPs of QAM reranking with OSPP.
    ``BEF.'' means the R-MAC results (in Tab. \ref{tab:exp:map_vgg19}) before QAM reranking.
    The cluster number is $K=25$ for FMP.
    %The results of ``FMP'' are from Tab. \ref{tab:fmp_different_k} where the cluster number is $K=25$.
    }
\label{tab:ospp_map}
\begin{tabular}{l|llll}
\hline
        & Ox5k  & Pa6k  & Sc6k  & INSTRE   \\ \hline
BEF.    & 0.695 & 0.838 & 0.490 & 0.415 \\
OSPP    & \textbf{0.736} & 0.842 & 0.505 & 0.436 \\
FMP     & 0.710 & \textbf{0.845} & \textbf{0.514} & \textbf{0.446}\\
\hline
\end{tabular}
\end{table}

\begin{table}[!t]
    \centering
    \caption{The computational cost per image on average for FMP and OSPP, including the feature dimensionality (DIM.), the memory footprints (MEM.) and the similarity computation time (TIME, in seconds).}
    \label{tab:exp:rerank_costs_vgg19}
    \begin{tabular}{l|ccc}
    \hline
         & DIM.            & MEM.   & TIME  \\ \hline
    FMP  & $25 \times 512$ & 51KB   & 0.016 \\
    OSPP & $23 \times 512$ & 47KB   & 0.014 \\
    \hline
    \end{tabular}
\end{table}

% \subsubsection{Discussion}
%     The above two sections show that the proposed QAM improves the retrieval performance by selecting the discriminative components of the target object that contribute most to the query-to-image similarity.
%     The effectiveness of different region descriptors shows that QAM provides a general framework to compare the similarity between the query and multiple regions.
%     In terms of memory footprints, the region descriptors (FMP) for one million images cost around 51GB, and the reranking time for top 100 images is 1.6 seconds on average per query.
%     Note that reranking with more images achieves higher mAP but takes more time as well.

%\begin{figure}[!t]
%    \centering
%    \includegraphics[width=.8\linewidth]{tab/0map_rerank}
%    \caption{The mAPs of VGG19 before and after QAM reranking on different datasets. 
%    The ``SPoC'' denotes results before reranking (Tab. \ref{tab:exp:map_vgg19}) and ``SPoC+FMP'' denotes the reranking results with FMP descriptors, and so on.}
%    \label{fig:exp:map_rerank_vgg19}
%\end{figure}

%%%%%%%%%%%%%%%% FMP, No ResNet  %%%%%%%%%%%%%%%%%%%%%%%%
\begin{savenotes}
\begin{table*}[!t]
\centering
\caption{
    Comparison with Existing Methods. 
    We report results for different retrieval methods with/without (bottom/top) post-processing.
    The ``Remark'' column shows the feature type used by each method.
    ``DIM.'' is the feature dimensionality.
    Methods with ``*'' are reported using our own implementations.
    }
\label{tab:exp:vs_soa}
\begin{tabular}{l|c|c|cccccc}
    \hline
                                    & Remark        & DIM. & Ox5k  & Ox105k & Pa6k  & Pa106k & Sc6k  & INSTRE \\ \hline
    J{\'{e}}gou and Zisserman \cite{Jegou:Tembed/2014}
                                    & SIFT        & 512  & 0.528 & 0.461  & -     & -      & -     & -      \\
    Babenko et al. \cite{Babenko:NeuralCodes/2014}       
                                    & CNN        & 512  & 0.557 & 0.522  & -     & -      & -     & -      \\
                                            % & 512  & 0.435 & 0.392  & -     & -      & -     & -      \\
    Razavian et al. \cite{Razavian:CnnINSBaseline/2015} 
                                    & CNN        & 256  & 0.533 & 0.489  & 0.670 & -      & 0.377 & -      \\
    SPoC* \cite{Babenko:spoc/2015}  & CNN        & 512  & 0.685 & 0.622  & 0.798 & 0.704  & 0.517 & 0.306  \\
    R-MAC* \cite{Tolias:rmac/2015}  & CNN        & 512  & 0.695 & 0.644  & 0.838 & 0.763  & 0.490 & 0.415  \\
    %SPoC \cite{Babenko:spoc/2015}           & 512  & 0.589 & 0.578  & 0.798* & 0.704*  & 0.517* & 0.306*  \\
    %R-MAC \cite{Tolias:rmac/2015}           & 512  & 0.669 & 0.616  & 0.830 & 0.757  & 0.490* & 0.415*  \\

    \hline
    J{\'{e}}gou et al. \cite{Jegou:HE/2010} & BoW & -    & 0.747 & -      & -     & -      & -     & 0.268\tablefootnote{This result is reported in \cite{Wang:instre/2015}}  \\
    Chum et al. \cite{Chum:qerevisit/2011}  & BoW & -    & 0.827 & 0.767  & 0.805 & 0.710  & -     & -      \\
    Mikul{\'{\i}}k et al. \cite{Mikulik:FineQuantization/2013} 
                                            & BoW & -    & \textbf{0.849} & \textbf{0.795} & 0.824  & 0.773 & - & -  \\
    Arandjelovic and Zisserman \cite{Arandjelovic:sculptures6k/2011}   
                                            & BoB & -    & -     & -      & -     & -      & 0.502 & -      \\
    Tolias et al. \cite{Tolias:rmac/2015}   & CNN & -    & 0.773 & 0.732  & 0.865 & 0.798  & -     & -      \\
    \textbf{Ours}                           & CNN & -    & 0.781 & 0.747  & \textbf{0.874} & \textbf{0.828}  & \textbf{0.608} & \textbf{0.592}  \\
    \hline
\end{tabular}
\end{table*}
\end{savenotes}

\subsection{Comparison With Existing Work}\label{sec:exp:vs_soa}
    Our complete retrieval pipeline consists of three steps: initial retrieval with R-MAC, QAM reranking with FMP, and QE in the end (see Sec. \ref{sec:cvx:pipeline}).
    Here we compare our complete retrieval method to the state of the art in Tab.~\ref{tab:exp:vs_soa}.
    
    The results are divided into two parts, with the above part being the retrieval methods using different compact representations but involving no post-processing, e.g., reranking, spatial verification, or QE, and the other part corresponding to methods that use post-processing strategies.
    
    From the first part we observe that when aggregated to comparable dimensionalities, CNN based representations can have better retrieval performance than SIFT based compact representations and among the CNN representations, R-MAC performs best since it considers both the multi-scale information and weak spatial information.

    % The first row of the table reports the performance of different compact representations, including the latest aggregation method for SIFT \cite{Jegou:Tembed/2014} and the ones using CNN features \cite{Babenko:NeuralCodes/2014,Razavian:CnnINSBaseline/2015,Babenko:spoc/2015,Tolias:rmac/2015}.
    % These methods do not include post-processing after initial retrieval (e.g., reranking, spatial verification, or QE), and serve as baselines for comparison.
    % Overall, the CNN based methods provide better performance.
    %For conventional local descriptors, J{\'{e}}gou and Zisserman \cite{Jegou:Tembed/2014} propose triangulation embedding and democratic aggregation to combine SIFT descriptors into a single vector representation.
    %For CNN features, Babenko et al. \cite{Babenko:NeuralCodes/2014} use the features from fully-connected layer, while Razavian et al. \cite{Razavian:CnnINSBaseline/2015}, SPoC \cite{Babenko:spoc/2015} and R-MAC \cite{Tolias:rmac/2015} use the features from the last convolutional layer.
    %These results serve as baselines for initial retrieval.
    %Both SPoC and R-MAC show good performance while being compact in dimensionality (with only 512), but R-MAC outperforms SPoC most of the time (except for Sculptures6k).
    
    The second part of Tab. \ref{tab:exp:vs_soa} reports methods with post-processing.
    J{\'{e}}gou et al. \cite{Jegou:HE/2010} improves the traditional BoW model with hamming embedding and weak geometric consistency constraints, but the result of this method on INSTRE reported in \cite{Wang:instre/2015} is far from satisfactory.
    Our method achieves the best performance on all datasets except for Oxford5k and Oxford105k, where the well developed BoW models \cite{Chum:qerevisit/2011,Mikulik:FineQuantization/2013} are still better. Higher mAPs (more than 0.9) on Oxford5k and Paris6k are reported in \cite{Arandjelovic:ThreeThings/2012} and \cite{Zhong:dsm/2015}. But since their methods learn the visual codebooks on the original datasets and hence are highly tailored, the results are not directly comparable.
    Compared to \cite{Arandjelovic:sculptures6k/2011} in which specialized representations, namely Bag-of-Boundaries (BoB), for smooth object are proposed, our method outperforms it by 10\% and has better generalization ability as well.
    Our main focus is on comparing to the latest work in \cite{Tolias:rmac/2015} since their representation is based on CFMs and we have similar retrieval pipelines. The most notable difference between their method and ours lies in the reranking strategy. For fair comparison, we use similar initial retrieval as \cite{Tolias:rmac/2015}. From the results we can see our method consistently has superior performance. Another point worth mentioning is that, as far as we know, our method achieves the best performance on the INSTRE dataset among the reported results which shows the robustness of our reranking method on dataset with cluttered backgrounds.

    % Our retrieval pipeline is similar to theirs but we use our proposed reranking method.
    % For objects with cluttered backgrounds (e.g., images in the INSTRE datasets), our method shows superior performance which shows that the proposed QAM is robust to distinguish the target object from distractors.
    Finally we show the search time of different components of the retrieval pipelines for each query and the memory consumption of the whole retrieval framework on different dataset sizes in Tab. \ref{tab:exp:total_time_mem}.
    % In the next section, we will show that the retrieval performance of our method will be further improved when equipped with better network. 
%%%%%%%%%%%%%%%%%  FMP's time & mem %%%%%%%%%%%%%%%%%%%%%%%%%%%%
\begin{table}[!t]
\centering
\caption{
    The searching time (TIME) and memory footprints (MEM.) on different dataset sizes (5k, 100k, 1M).
    The search time per query includes initial search (INIT., excluding query feature extraction), reranking (QAM) and query expansion (QE, the same as initial search).
    The memory footprints required by all database images include global features (R-MAC) and multi-region features (FMP).
    The total amount of costs are summarized in the ``tot.'' columns.
}
\label{tab:exp:total_time_mem}
\scalebox{0.85}{
\begin{tabular}{c|llcl|r|r|r}
\hline
\multicolumn{1}{l|}{\multirow{2}{*}{}} & \multicolumn{4}{c|}{TIME (seconds)}                                                                                                                                                              & \multicolumn{3}{c}{MEM.}                                                          \\ \cline{2-8} 
\multicolumn{1}{l|}{}                  & \multicolumn{1}{c|}{INIT.} & \multicolumn{1}{c|}{QAM}   & \multicolumn{1}{c|}{QE}                                                                        & \multicolumn{1}{c|}{tot.} & \multicolumn{1}{c|}{R-MAC} & \multicolumn{1}{c|}{FMP} & \multicolumn{1}{c}{tot.} \\ \hline
5K                                     & \multicolumn{1}{l|}{0.001}  & \multicolumn{1}{l|}{1.161} & \multicolumn{1}{c|}{\multirow{3}{*}{-}} & 1.163                     & 10MB                        & 256MB                     & 266MB                    \\
100K                                   & \multicolumn{1}{l|}{0.017}  & \multicolumn{1}{l|}{1.160} & \multicolumn{1}{c|}{}                                                                          & 1.194                     & 200MB                       & 5.1GB                     & 5.3GB                    \\
1M                                     & \multicolumn{1}{l|}{0.127}  & \multicolumn{1}{l|}{1.374} & \multicolumn{1}{c|}{}                                                                          & 1.628                     & 2GB                         & 51.2GB                      & 53.2GB                     \\ \hline
\end{tabular}
}
\end{table}

\subsection{Performance with Deeper Networks}\label{sec:exp_deeper_nets}
    Recent works discover that deeper neural networks can lead to significant improvement on tasks like image classification and object detection. Here we explore two of these deeper networks on instance retrieval. One is GoogLeNet \cite{Szegedy:GoogLeNet/2015} which is a 22-layer network incorporating the ``Inception'' modules and the other one is a recently proposed 152-layer convolutional neural network \cite{He:ResNet/2015}, named deep residual net (ResNet152).
    % However, the convolutional layers are still the common building blocks for both GoogLeNet and ResNet152.
    % Therefore, our interest is to investigate whether the CFMs from these deeper networks are still effective for instance retrieval.
    %Therefore, our interest is to investigate whether the feature patterns of CFMs are still persist in these deeper networks.
    
\subsubsection{Feature Patterns}    
    To intuitively understand the CFMs of these two networks, we first visualize the feature patterns of the CFMs of these two networks.
    Specifically, for each layer's CFMs, we calculate the $\ell_1$-norm for all the local descriptors ($||\mathbf{x}_i||_1$) and normalize them by the maximum norm value in this layer.
    The heat maps of these norm values indicate the activation degree of different image parts. 
    From the better explored AlexNet or VGG net we know that generally the CFMs of lower layers correspond to some primary patterns like edges, and CFMs of higher layers capture some object-related semantics. And the heat maps tend to be spatially sparse. 
    Here we observe some difference for the CFMs of GoogLeNet and ResNet152, the visualizations of which on the first example image from Fig. \ref{fig:example_imgs} are shown in Fig. \ref{fig:cfm_patterns_deeper_nets}.
    
    For GoogLeNet, the structures of the objects persist in CFMs from Inception 4a to 4e layer where the activated patterns are sparse and correspond to some meaningful object parts but the structure information tends to be lost in higher layers (5a and 5b).
    For ResNet152, the sparse property does not hold in lower layers (conv1 to res4) but persist in higher layers (res5a-res5c). Since the activated pattern plays an important role in CFMs aggregation, we attempt to unveil some relationship between the retrieval performance of CFMs and their corresponding feature patterns.

    % ResNet152's CFMs show very different sparsity patterns in different layers.
    % We infer that it is caused by the batch normalization \cite{Ioffe:BN/2015}, which ensures the forward propagated signals to have non-zero variances.
    % Interestingly, the spatial patterns and sparsity still persist in very deep layers (res5a - res5c).

\begin{savenotes}
\begin{figure*}[!t]
\centering
\includegraphics[width=.8\linewidth]{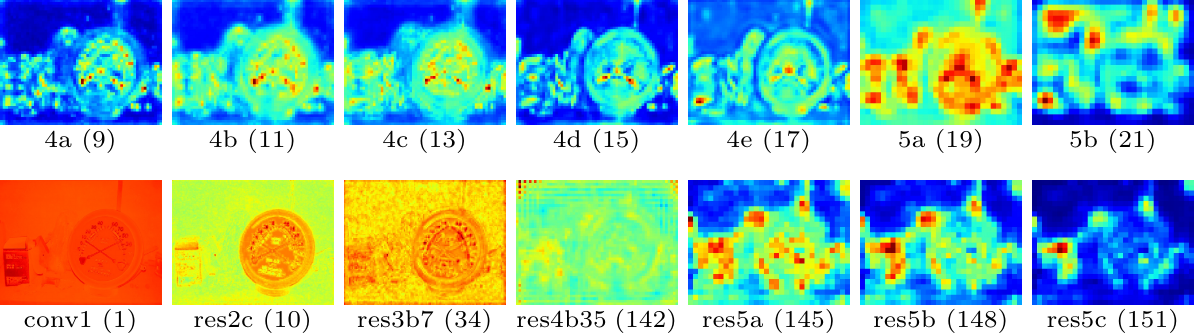}
\caption{
    The heat map of local descriptors' $\ell_1$-norms from GoogLeNet's Inception layers (top) and  ResNet152's residual layers (bottom). 
    Below are the layers' names and their depths (e.g., ``4a (9)'' means the ``Inception 3a'' layer and its depth is 9 in GoogLeNet). The layers' names are consistent with the network structures defined in their Caffe ``.prototxt'' files.
}
\label{fig:cfm_patterns_deeper_nets}
\end{figure*}
\end{savenotes}
%    \begin{figure*}[!t]
%        \centering
%        \includegraphics[width=.8\linewidth]{fig/vis_normxy_googlenet}
%        \caption{The heat map of local descriptors' $\ell_1$-norms from GoogLeNet's Inception modules \cite{Szegedy:GoogLeNet/2015}. Below are the layers' names (e.g., ``3a'' means the ``Inception 3a'' layer).}
%        \label{fig:vis_normxy_googlenet}
%    \end{figure*}
%    
%    \begin{figure*}[!t]
%        \centering
%        \includegraphics[width=.8\linewidth]{fig/vis_normxy_res152}
%        \caption{The heat map of local descriptors' $\ell_1$-norms from ResNet152's residual modules \cite{He:ResNet/2015}. Below are the layers' name (e.g., ``res2c'' means the ``residual 2c'' layer). These namings follow the authors' publicly released Caffe model.}
%        \label{fig:vis_normxy_res152}
%    \end{figure*}

\subsubsection{Choosing the Right Layer}
\begin{figure*}[!t]
    \centering
    \includegraphics[width=.9\linewidth]{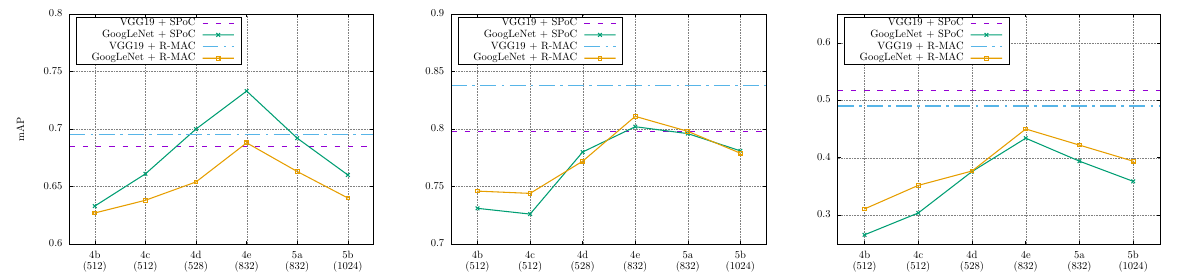}
    \caption{
    The mAPs of GoogLeNet's different layers on Oxford5k, Paris6k and Sculptures6k (from left to right). 
    The ``VGG19 + SPoC'' and ``VGG19 + R-MAC'' are the baseline results of VGG19 in Tab. \ref{tab:exp:map_vgg19}.   
    The horizontal axis gives the layers' names and their feature dimensionalities after aggregation (e.g., ``4b (512)'' means the ``Inception 4b'' layer and the dimensionality of its aggregated feature is 512).
    }
    \label{fig:exp:map_googlenet}
\end{figure*}

    Here we evaluate the retrieval performance of CFMs from different layers using different global aggregation methods (Sec. \ref{sec:exp:settings_global_features}).
    For GoogLeNet, we select the last six Inception modules (Inception 4b to Inception 5b) for testing.
    Fig. \ref{fig:exp:map_googlenet} shows the retrieval performance of different layers.
    As we can see, initially the mAPs on all datasets are improved from lower to higher layers (from 4b to 4e), but the last two Inception layers (5a and 5b) have inferior performance. The reason of the performance degradation may lie in our aformentioned analysis that the spatial structure and sparsity are weakened in the two higher layers.
    On Paris6k and Sculptures6k, GoogLeNet is outperformed by VGG19 and on Oxford5k GoogleNet performs better.
    % Besides, the aggregated feature from the best layer (Inception 4e) has higher dimensionality than the one in VGG19 (832 vs 512).
     
\begin{figure*}[!t]
    \centering
    \includegraphics[width=.9\linewidth]{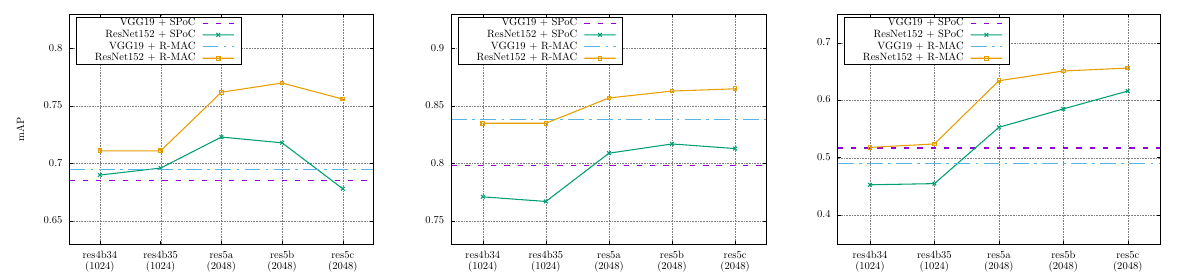}
    \caption{
    The mAPs of ResNet152's different layers on Oxford5k, Paris6k and Sculptures6k (from left to right).
    Similar to Fig. \ref{fig:exp:map_googlenet}, below are the layers' names and their dimensionalities (e.g., ``res4b34 (1024)'' means the ``residual 4b34'' layer and the dimensionality of its aggregated feature is 1024).
    }
    \label{fig:exp:map_res152}
\end{figure*}

    For ResNet152, we select the last five residual modules (res4b34 to res5c) for testing as shown in Fig. \ref{fig:exp:map_res152}. From this figure we observe that generally higher layers perform better and the best performed setting of ResNet152, i.e., res5b with R-MAC, can signifiantly outperform VGG19.
    However, the improvements are bought at the price of more memory and computational cost: the dimensionality of ResNet152's best performed layer, res5b, is 2,048, which is four times higher than that in VGG19.
    
\subsubsection{Final Integration}
    Finally, we test our methods with CFMs from the res5b layer of ResNet152.
    The whole retrieval pipeline is still the same: initial retrieval with R-MAC, followed by QAM re-ranking with OSPP, and finally QE.
    Note that OSPP is applied here instead of FMP because it takes too much time to downsize the FMP features from $\mathbb{R}^{2048 \times 2048}$ to $\mathbb{R}^{25 \times 2048}$ via clustering for each image.
    The results are shown in Tab.~\ref{tab:exp:map_res152}.
    We can see the ResNet152 can significantly outperform VGG19 on all the datasets.
    The results demonstrate the effectiveness of deeper network in instance retrieval.
    However, these improvements are brought at the price of more memory and computational cost.

%%%%%%% vgg19 FMP vs resnet ospp %%%%%%%
\begin{savenotes}
\begin{table}[!t]
\centering
%\small
\caption{
    The mAPs of our retrieval framework with different CFMs of different networks.
}
\label{tab:exp:map_res152}
\scalebox{0.8}{
\begin{tabular}{l|cccccc}
    \hline
                                    & Ox5k  & Ox105k & Pa6k  & Pa106k & Sc6k  & INSTRE \\ \hline
    VGG19                           & 0.781 & 0.747  & 0.874 & 0.828  & 0.608 & 0.592  \\
    ResNet152                       & \textbf{0.828} & \textbf{0.814}  & \textbf{0.893} & \textbf{0.858}  & \textbf{0.712} & \textbf{0.670}  \\
    \hline
\end{tabular}
}
\end{table}
\end{savenotes}

\section{Conclusion}\label{sec:conclusion}
    In this work, we propose a reranking algorithm, namely query adaptive matching, for instance retrieval using convolutional feature maps. The key idea is to represent a candidate database image by a set of base regions and generate a target object focused representation resorting to the similarity to the query image. We formulate the similarity matching as well as the region merging process as an optimization problem which can be solved efficiently. Apart from this general framework, we propose two practical ways to generate the base regions. The experiments on several instance retrieval datasets demonstrate the effectiveness of the proposed reranking method. 
    
    % Our method formulates the similarity matching between a query representation and a candidate database image  as an optimization problem and effectively merges multiple base regions into the region-of-interest w.r.t. the target object.
    % We also propose two different ways to generate the base regions.
    % Experiments show that the proposed reranking methods can achieve promising results which outperform the state of the art.
    % Besides that, we evaluate our methods with deep CNN models and demonstrate further improvement.
    
%%%%%%%%%%% MM 2016 %%%%%%%%%%%%%%%
\bibliographystyle{abbrv}
\bibliography{qam}
%\bibliography{CNN-Locality}

\end{document}